\documentclass[10pt,twocolumn,letterpaper]{article}

\usepackage{cvpr}
\usepackage{times}
\usepackage{epsfig}
\usepackage{graphicx}
\usepackage{amsmath}
\usepackage{amssymb}
\usepackage{booktabs,subcaption,amsfonts,dcolumn}
\usepackage{float}
\usepackage{comment}
\usepackage{array}
\usepackage{amsmath}
\usepackage{amssymb}
\usepackage{epstopdf}
\usepackage{multirow}
\usepackage{array}
\usepackage[linesnumbered,ruled]{algorithm2e}
\newcolumntype{L}[1]{>{\raggedright\let\newline\\\arraybackslash\hspace{0pt}}m{#1}}
\newcolumntype{C}[1]{>{\centering\let\newline\\\arraybackslash\hspace{0pt}}m{#1}}
\newcolumntype{R}[1]{>{\raggedleft\let\newline\\\arraybackslash\hspace{0pt}}m{#1}}

\usepackage[pagebackref=true,breaklinks=true,letterpaper=true,colorlinks,bookmarks=false]{hyperref}

\cvprfinalcopy 

\def\cvprPaperID{****} 

\ifcvprfinal\fi
\begin{document}

\title{Thickened 2D Networks for Efficient 3D Medical Image Segmentation}

\author{
Qihang Yu\textsuperscript{1}~~~~
Yingda Xia\textsuperscript{1}~~~~
Lingxi Xie\textsuperscript{2}~~~~
Elliot K. Fishman\textsuperscript{3}~~~~
Alan L. Yuille\textsuperscript{1}~~~~ \vspace{.3em}\\
\textsuperscript{1} The Johns Hopkins University \qquad
\textsuperscript{2} Noah's Ark Lab, Huawei Inc. \\
\textsuperscript{3} The Johns Hopkins Medical Institute\\
{\tt\small \{yucornetto, 198808xc, alan.l.yuille\}@gmail.com}\quad{\tt\small yxia25@jhu.edu}\quad{\tt\small efishman@jhmi.edu}
}

\maketitle

\begin{abstract}
There has been a debate in 3D medical image segmentation on whether to use 2D or 3D networks, where both pipelines have advantages and disadvantages. 2D methods enjoy a low inference time and greater transfer-ability while 3D methods are superior in performance for hard targets requiring contextual information. This paper investigates efficient 3D segmentation from another perspective, which uses 2D networks to mimic 3D segmentation. To compensate the lack of contextual information in 2D manner, we propose to thicken the 2D network inputs by feeding multiple slices as multiple channels into 2D networks and thus 3D contextual information is incorporated. We also put forward to use early-stage multiplexing and slice sensitive attention vto solve the confusion problem of information loss which occurs when 2D networks face thickened inputs. With this design, we achieve a higher performance while maintaining a lower inference latency on a few abdominal organs from CT scans, in particular when the organ has a peculiar 3D shape and thus strongly requires contextual information, demonstrating our method's effectiveness and ability in capturing 3D information. We also point out that ``thickened" 2D inputs pave a new method of 3D segmentation, and look forward to more efforts in this direction. Experiments on segmenting a few abdominal targets in particular blood vessels which require strong 3D contexts demonstrate the advantages of our approach.
\end{abstract}

\newcommand{\figurewidth}{8cm}
\begin{figure}[!t]
\centering
\includegraphics[width=\figurewidth]{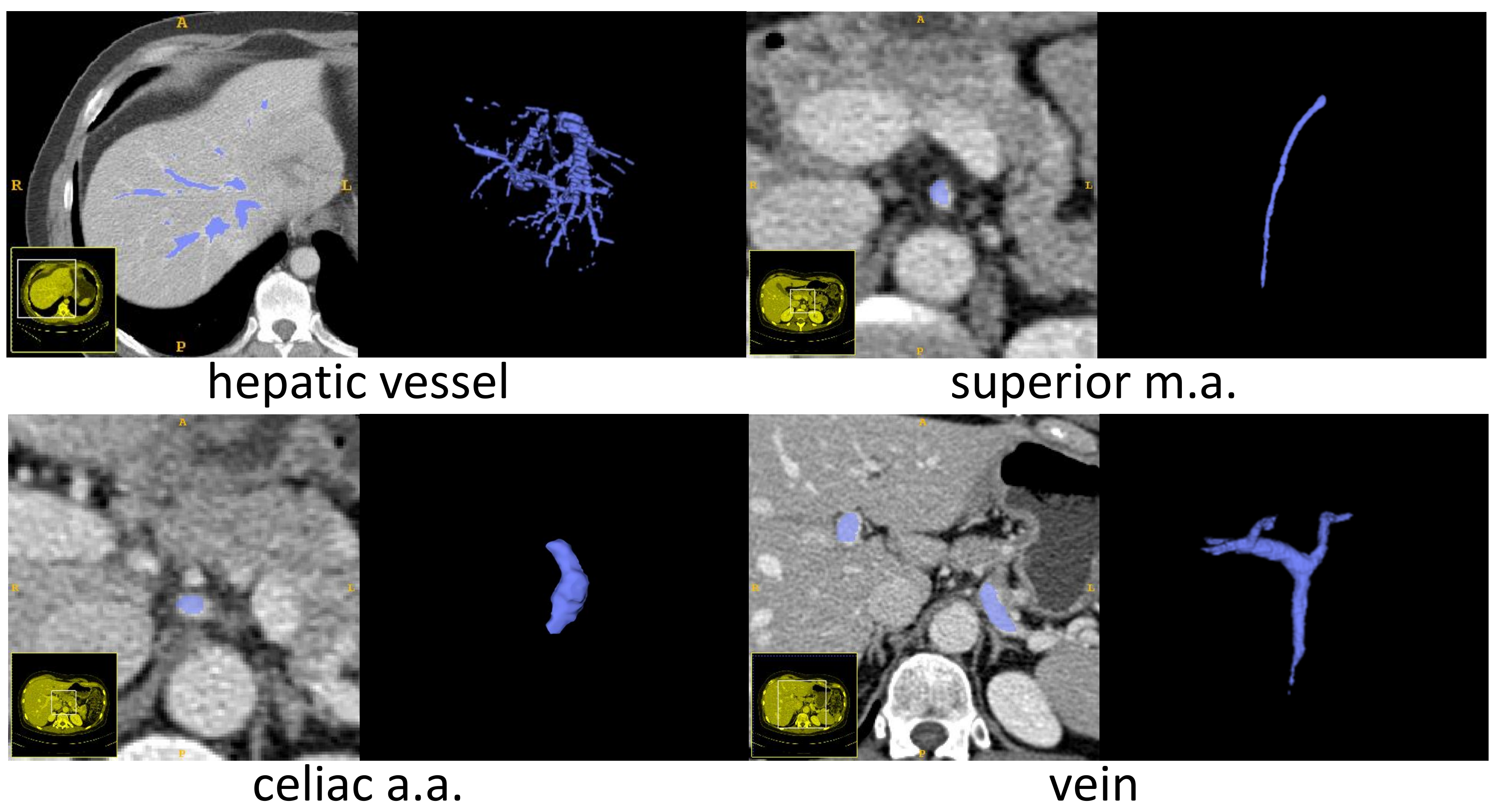}

\caption{Examples for hepatic vessel, superior m.a., celiac a.a., vein in the CT scan respectively. Image is shown with 2D label denoted by blue. Vessels are small, and the shape features in spatial continuity, so it is difficult to segment vessels for a single-view based 2D network. (Best viewed in color.)}
\vspace{-0.4cm}
\label{Fig:Motivation}
\end{figure}

\section{Introduction}

 \begin{figure*}[!t]
\centering
\includegraphics[width=14.5cm]{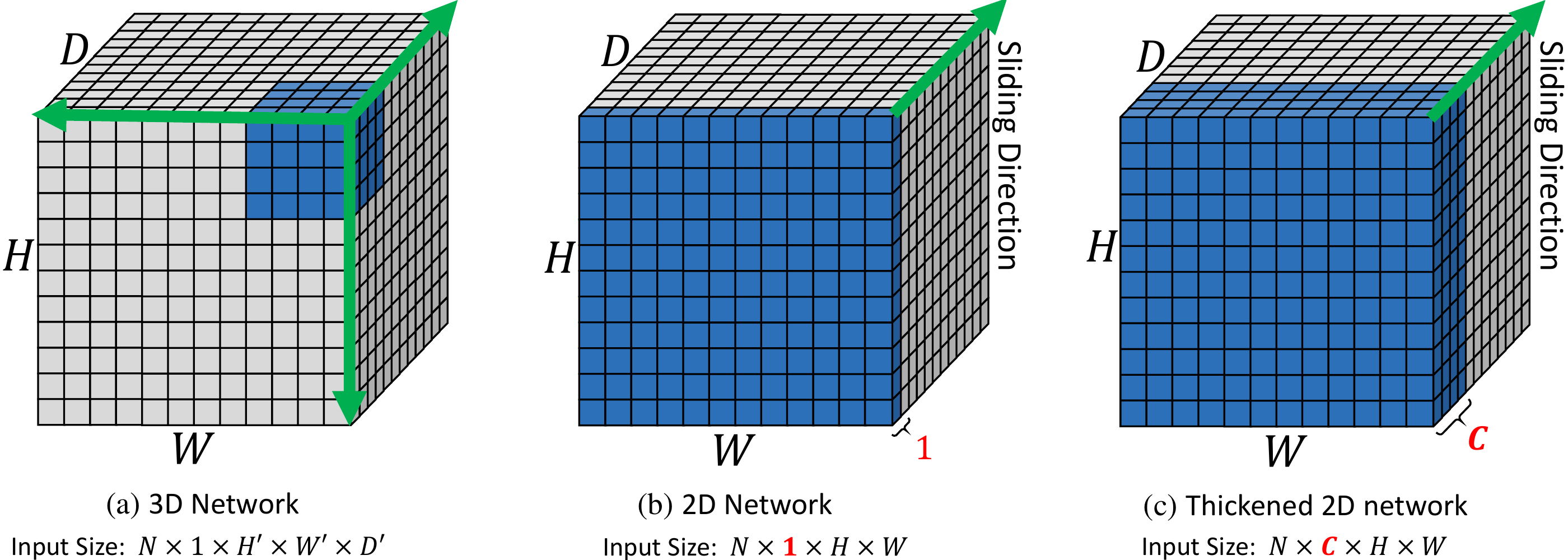}

\caption{Toy examples of different methods dealing with a 3D volume, including (a) 3D network (input patch size $1\times H'\times W'\times D'$), (b) 2D network (input single slice $1\times H\times W$), (c) thickened 2D network (input $C$-slice $C\times H\times W$). At training, 2D, 3D, T2D network is trained with single full-size slices, cropped patches, and thickened full-size slices respectively. At testing stage, 2D and T2D slide along single axis and stack the slice-prediction into a volume prediction, while 3D slides along all three axes in a sliding window manner. (Best viewed in color.)}
\vspace{-0.2cm}
\label{Fig:ToyExample}
\end{figure*}

Medical image segmentation is an important prerequisite of computer-assisted diagnosis (CAD) which implies a wide range of clinical applications. In this paper, we focus on organ segmentation with 3D medical images, such as Computed Tomography (CT) and Magnetic Resonance Imaging (MRI). To deal with volumetric input, two main flowcharts exist. The first method borrows the idea from natural image segmentation, cutting the 3D volume into 2D slices, and training a 2D network which deals with each slice individually or sequentially. Another way is to crop patches from the volume, and train a 3D network to deal with volumetric patches directly. Afterwards, both methods test the original volume in a sliding window manner.

Both of these two methods have their own advantages and disadvantages. A 2D network, since it takes a full-size slice as input and thus only needs to slide along single axis, has a much lighter computation and higher inference speed, but suffers from the lack of information of relationship among slices (see Figure~\ref{Fig:Motivation}). For mainstream patch-based 3D approach, in opposite, has the perception of 3D contexts, but also suffers from several weaknesses: 
(i) patch-based method has a limited receptive field, and makes the model easier to be confused;
(ii) the lack of pre-trained models which makes the training process unstable; and (iii) in the testing stage, patch-based method requires sliding along all three axes, resulting in a high inference latency.

This paper presents a novel framework which {\bf uses a 2D network to mimic 3D segmentation by thickening 2D inputs}. Therefore the model enjoys both the light computation of 2D method and ability to capture contextual information of 3D method. A toy example illustrating the differences among 2D, 3D, and thickened 2D (T2D) at training and testing is shown in Figure~\ref{Fig:ToyExample}. Our idea is motivated by a few prior works for pseudo-3D segmentation~\cite{Yu_2018_CVPR,zhou2017fixed}, in which three neighboring 2D slices are stacked during training and testing, so that the 2D network sees a small range of 3D contexts. However, these approaches still produce unsatisfying results for the blood vessels, because recognizing these targets requires even stronger 3D contexts -- see Figure~\ref{Fig:Motivation} for examples. To deal with this problem, a natural choice is to continue thickening the input of 2D networks, allowing richer 3D contexts to be seen. 

However, directly increasing slice thickness reaches a plateau quickly at 5 or 6 slices, and adding more slices to 2D networks causes accuracy drop. The essence lies in information loss. Technically, for a 2D network, information from all input slices are mixed together into channel dimension, and, without a specifically designed scheme, it is difficult to discriminate them from each other. For a 2D network with thickened input, information from different slices are fused in the first convolution. Since background noises are not filtered, mixing slices together makes it hard to pick up useful information for distinguishing each slice. We argue that this early fusion (information from all slices are fused in the first convolution) is what leads to the information loss and the worse performance with thickened inputs.

Therefore, we propose two solutions based on a 2D segmentation backbone to address this problem. (1) \textbf{Early-Stage Multiplexing (ESM)}: we postpone the stage that these information are put together. The backbone is divided into two parts, and the thickened input is divided into several mini-groups. We multiplex first part of the backbone to deal with each mini-group individually, while in the second part the information from different slices are fused. (2) \textbf{Slice-Sensitive Attention (SSA)}: to improve the discriminative power of the fused feature maps, we introduce slice-sensitive attention between the pre-fusion stage and the decision stage. The options of fusing these two sources of information are also studied in an empirical manner.

Experiments are conducted on several abdominal organs individually, including two regular organs and three blood vessels in our own dataset, and the hepatic vessels in the Medical Segmentation Decathlon (MSD)~\cite{simpson2019large} dataset. Our approach achieves a better performance compared with popular 3D models in terms of dice score and also has a lower inference latency. We also prove that our method improves the ability to distinguish neighboring slices with a designed metric.

The remainder of this paper is organized as follows. Section~\ref{RelatedWork} briefly reviews related work, and Section~\ref{Approach} presents our approach. After experiments are shown in Section~\ref{Experiments}, we conclude this work in Section~\ref{Conclusions}.

\section{Related Work}
\label{RelatedWork}

Computer aided diagnosis (CAD) is a research area aiming at helping human doctors in clinics. Recently a lot of CAD approaches are based on medical imaging analysis to get accurate descriptions of the scanned organs, soft tissues, $etc.$ One topic with great importance in this area is object segmentation, $i.e$, determining which voxels belong to the target in 3D data. In natural image segmentation, conventional methods based on graph~\cite{ali2007graph, papon2013voxel} or handcrafted local features~\cite{wang2014geodesic} have been gradually replaced by techniques from deep learning, which could produce higher segmentation accuracy~\cite{chen2018deeplab,long2015fully,zhao2017pyramid}. And it has been proved successful not only in the natural image area but also medical image area~\cite{milletari2016v,ronneberger2015u}, outperforming conventional approaches, $e.g.$, when segmenting the brain~\cite{chen2018voxresnet, kamnitsas2017efficient}, the liver~\cite{dou20173d,li2017h}, the lung~\cite{hu2001automatic, jin2018ct}, or the pancreas~\cite{chu2013multi,roth2016spatial}.

These deep learning methods in medical image segmentation can be classified into the following types according to their way to deal with 3D volumetric data:

Researchers taking \textbf{2D method} cut each 3D volume into 2D slices, and train a 2D network to process each of them individually~\cite{oktay2018attention,ronneberger2015u}. Such methods often suffer from missing 3D contextual information, for which various techniques are adopted, such as using 2.5D data (stacking a few 2D images as different input channels)~\cite{roth2015deeporgan,roth2016spatial}, training deep networks from different viewpoints and fusing multi-view information at the final stage~\cite{xia20183d,xia2018bridging}, and applying a recurrent network to process sequential data~\cite{cai2016pancreas,chen2016combining}.

A \textbf{3D workflow}, researchers directly train a 3D network to deal with the volumetric data~\cite{cciccek20163d,milletari2016v}. These approaches, while being able to see more contextual information, often require much larger memory consumption. So some existing methods work on small patches and fuse the outputs of all patches~\cite{havaei2017brain, zhu20173d}, or down-sample the whole volume into a lower resolution~\cite{gibson2018densevnet,milletari2016v}. In addition, unlike 2D networks that can borrow pre-trained models from natural image datasets, 3D networks need to be trained from scratch, which means it could suffer from unstable convergence properties~\cite{tajbakhsh2016convolutional}. 
A discussion on 2D vs. 3D models for medical imaging segmentation is available in~\cite{lai2015deep}.

Some works also explore the way to find a way to \textbf{combine 2D and 3D methods}.~\cite{xia2018bridging} proposes to use a 3D network to fuse 2D predictions from different viewpoints.~\cite{ni2018elastic} tries to use a 2D network which predicts the distance from voxel to organ boundary, and afterwards, 3D reconstruction is applied to generate a final dense prediction. In~\cite{liu20183d}, a hybrid segmentation network consisting of 2D encoder and 3D decoder is designed, thus the network benefits from 2D pre-trained weight and 3D contextual information.

In this paper, we provide an alternative way to incorporate 3D contextual information into 2D networks. Our method differs from what mentioned above by (1) the third dimension is incorporated into channel dimension, thus no 3D operation or module is introduced and (2) the model is still in 2D manner thus enjoy a light computation cost at inference.

\begin{figure*}[t]
\begin{center}
  \includegraphics[width=1.0\linewidth]{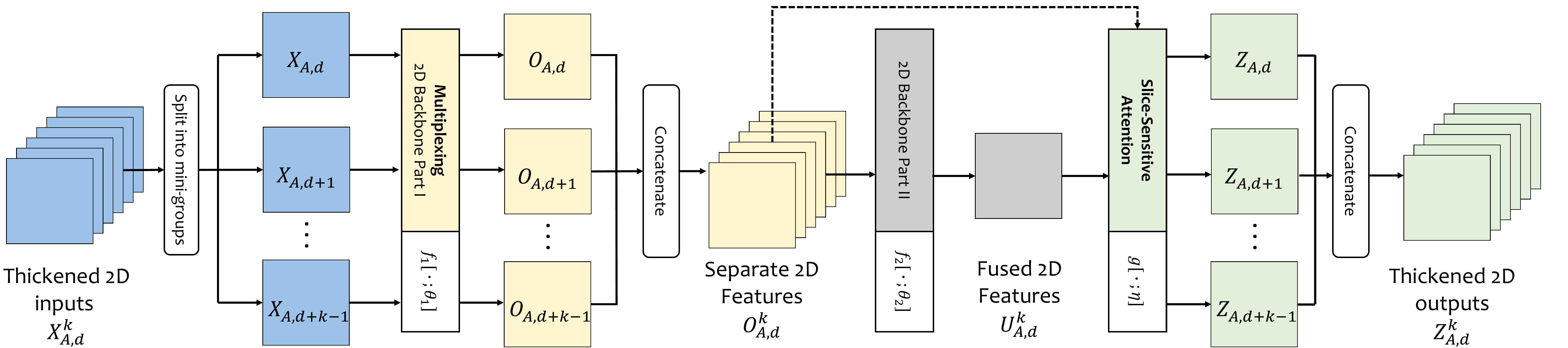}
\end{center}
  \caption{The proposed architecture for processing thickened inputs. It divides the backbone into two parts. Information is propagated in the first part and fused in the second one. By multiplexing the first part of the backbone, this method only requires a small number of extra parameters. Different feature maps are denoted in different colors. (Best viewed in color.)}
\label{fig:network_arch}
\end{figure*}

\section{Our Approach}
\label{Approach}

\subsection{Problem Statement and 2D/3D Baselines}

\subsubsection{Problem Statement}
In this task, a CT scan is regarded as a 3D volume $\mathbf{X}$ of size $H \times W \times D$, and each voxel $X_{h,w,d}$ indicates the intensity at the specified position measured by the Haunsfield unit (HU). It is annotated with a binary ground-truth segmentation $\mathbf{Y}$ where $y_{i}$ = 1 means a foreground voxel. Suppose our model predicts a volume $\mathbf{Z}$, and $\mathcal{Y}=\{\ (h,w,d)\ |\ Y_{h,w,d}=1\}$ and $\mathcal{Z}=\{\ (h,w,d)\ |\ Z_{h,w,d}=1\}$ are the foreground voxels in the ground-truth and prediction, respectively. The segmentation accuracy is evaluated using the Dice-S\o rensen coefficient (DSC): $\mathrm{DSC}$ $(\mathcal{Y},\mathcal{Z}) = \frac{2\times|\mathcal{Y}\cap \mathcal{Z}|}{|\mathcal{Y}|+|\mathcal{Z}|}$, which has a range of $[0, 1]$ with 1 implying a perfect prediction.

\subsubsection{2D and 3D Baselines}
2D and 3D models are most common methods in nowadays medical image segmentation. \textbf{2D segmentation networks} deal with 3D volume by cutting them into stacked 2D slices firstly. Usually, the slicing procedure is based on three axes, the \textit{coronal}, \textit{sagittal}, and \textit{axial}. On each axis, an individual 2D deep network is trained. Final prediction is obtained by stacking 2D slices predictions, and the contextual information is added in a post-processing manner through fusing three-viewpoints results. In contrary, \textbf{3D segmentation networks} deal with 3D volume directly. Due to limitation of GPU memory, the mainstream 3D methods usually crop the volume into smaller patches instead of taking the entire volume as input, especially when data is in high resolution. In the testing stage, a sliding window of the same size is moved regularly along three axes, and prediction at each voxel is averaged over all predictions it gets.

A problem for \textbf{2D segmentation networks} is the lack of 3D contextual information. Although the final three-axes fusion makes compensation to some degree, it is still single-slice-based when training and testing. It is enough for some easy and large-scale organ like kidney, liver, and spleen, yet it works badly when facing more challenging tasks. \textbf{3D segmentation networks} also suffer from several deficits: (i) need to be trained from scratch, which could bring unstable convergence properties issues; (ii) sliding window manner is time-consuming and patch-based method limits the receptive field the network can see, thus it cannot get enough global information for an individual slice and is easy to get confused.

The drawbacks lie in 2D and 3D methods motivate us to propose thickened 2D network, where we feed multiple slices as a multiple channels into 2D network. By incorporating contextual information into 2D network, we expect the method to be more powerful than pure 2D methods and more robust and faster than pure 3D methods.

\subsection{Thickened 2D Inputs}
\subsubsection{Re-visiting 2D Segmentation}
Suppose the input is a 3D volume $\mathbf{X}$ of size $H \times W \times D$. When training single-slice based 2D deep networks for 3D segmentation, each 3D volume is sliced along three axes, the \textit{coronal}, \textit{sagittal}, and \textit{axial}. We denote these 2D slices with $\mathbf{X}_{\mathrm{C},h}$ ($h=1,2,...,H$), $\mathbf{X}_{\mathrm{S},w}$ ($w=1,2,...,W$) and $\mathbf{X}_{\mathrm{A},d}$ ($d=1,2,...,D$), where the subscripts $\mathrm{C}$, $\mathrm{S}$ and $\mathrm{A}$ stand for \textit{coronal}, \textit{sagittal}, and \textit{axial}, respectively. The models trained on each axis are denoted by $\mathbb{M}_{\mathrm{C}}$, $\mathbb{M}_{\mathrm{S}}$, and $\mathbb{M}_{\mathrm{A}}$, respectively. We consider a 2D slice along the \textit{axial} view, denoted by $\mathbf{X}_{\mathrm{A},d}$. Our goal is to infer a binary segmentation mask $\mathbf{Z}_{\mathrm{A},d}$ of the same shape. Usually, this is achieved by first computing a probability map $\mathbf{P}_{\mathrm{A},d} = \textbf{f}[\mathbf{X}_{\mathrm{A},d};\theta]$, where $\textbf{f}[\cdot;\theta]$ is a deep segmentation network with $\theta$ being network parameters, and afterwards, all predictions on single slice are concatenated into a 3D volume $\mathbf{P}_{\mathrm{A}}=[\mathbf{P}_{\mathrm{A},1},\mathbf{P}_{\mathrm{\mathrm{A}},2},\dots,\mathbf{P}_{\mathrm{A},D}]$. The final prediction are fused on three axes, $\mathbf{P} = \mathbb{F}(\mathbf{P}_{\mathrm{C}}, \mathbf{P}_{\mathrm{S}}, \mathbf{P}_{\mathrm{A}})$, where $\mathbb{F}$ is a fusion function, and $\mathbf{P}$ is binarized into $\mathbf{Z}$ using a fixed threshold, say 0.5, $i.e.$, $\mathbf{Z}=\mathbb{I}\left[\mathbf{P} \geqslant 0.5\right]$.

The prediction of 2D method is only based on single slice, which results in a lack of contextual information. Although 3D information is compensated in a post-processing manner through fusing predictions from three viewpoints, it is not enough to produce satisfying results on some hard target. Therefore, we put forward with incorporating contextual information into 2D network by thickening the 2D inputs.

Without loss of generality, the input is a k-slice group $\mathbf{X}_{\mathrm{A},d}^{k} = [\mathbf{X}_{\mathrm{A},d},\mathbf{X}_{\mathrm{A},d+1},\dots,\mathbf{X}_{\mathrm{A},d+k-1}]$, which are the slices starting from $d$ to $d+k-1$. The model also outputs a corresponding k-slice prediction $\mathbf{P}_{\mathrm{A},d}^{k}=\textbf{f}[\mathbf{X}_{\mathrm{A},d}^{k};\theta]$, where $\mathbf{P}_{\mathrm{A},d}^{k}=[\mathbf{P}_{\mathrm{A},d},\mathbf{P}_{\mathrm{A},d+1},\dots,\mathbf{P}_{\mathrm{A},d+k-1}]$. $\mathbf{P}_{\mathrm{A}}$ is obtained with a re-group function $\mathcal{G}$, i.e. $\mathbf{P}_{\mathrm{A}}=\mathcal{G}[\mathbf{P}_{\mathrm{A},1}^{k},\mathbf{P}_{\mathrm{A},2}^k,\dots,\mathbf{P}_{\mathrm{A},D-k+1}^{k}]$. In this paper, the re-group function $\mathcal{G}$ is assembling the groups and averaging on overlapped slices.

\vspace{-0.2cm}
\subsubsection{Information Loss}
Despite that increasing slice thickness to a moderate degree boosts performance, we observe that increasing thickness to a larger number results in performance drop. We notice that the reason is the information loss brought by 2D convolution operation which fuses different slices. A 2D convolution can be regarded as weighted-sum of all input channels for each output channel, so there is no special connection between specific output channel and corresponding input channel, which leads to a confusion of intra-slice information and inter-slice information.
Say $\textbf{R}^{C}$ is a feature map with $C$ channels, in a typical ResNet network, the feature map mapping is $\textbf{R}^{k}\rightarrow \textbf{R}^{64}\rightarrow \dots \rightarrow  \textbf{R}^{2048}\rightarrow \textbf{R}^{256}\rightarrow \textbf{R}^{k}$. For the input $\textbf{R}^{k}$ and output $\textbf{R}^{k}$, their channels should have one-to-one relationship in a $k$-slice medical image segmentation. When $k$ is small, say 1 or 3, it is easy for the network to figure out the mapping relations. But when $k$ is large, say 12, the network can be confused after so many 2D convolutions, which result in a loss of slice-sensitive information, and this information loss leads to confusion and a worse result.

\subsection{Adapting 2D Networks for Thickened Input}
The key of alleviating such information loss with thickened 2D inputs is to postpone the stage that information from multiple slices is fused and introduce slice-sensitive features. For this purpose, we propose early-stage multiplexing (ESM) and slice-sensitive attention (SSA) to address information loss for thickened inputs.

\vspace{-0.2cm}
\subsubsection{Early-Stage Multiplexing}
Firstly, one way to address the information loss is to encode every slice into a same feature space before fusion, which reduces the variance among different slices and also filters out large amount of useless background information. Specifically, we use multiple small-thickness groups instead of one large-thickness group. By multiplexing part of the backbone for each mini-group, the fusion stage is postponed. Without loss of generality, we divide the original 2D segmentation backbone $\textbf{f}[\cdot;\theta]$ into two parts, $\textbf{f}_{1}[\cdot;\theta_{1}]$ and $\textbf{f}_{2}[\cdot;\theta_{2}]$. In the first part, each mini-group forward propagates individually, therefore the intra-slice information is learned for each slice. Then in the remaining part, mini-groups get fused as one unity and inter-slice information is explored. With this modification, segmentation procedure on mini-groups with $k$-slice input becomes:

\begin{equation}
\small
\begin{aligned}
    &\mathbf{X}_{\mathrm{A},d}^{k} = [\mathbf{X}_{\mathrm{A},d},\mathbf{X}_{\mathrm{A},d+1},\dots,\mathbf{X}_{\mathrm{A},d+k-1}],\\
    &\mathbf{O}_{\mathrm{A},d} = \textbf{f}_{1}[\mathbf{X}_{\mathrm{A},d};\theta_{1}],\\
    &\mathbf{O}_{\mathrm{A},d}^{k} = [\mathbf{O}_{\mathrm{A},d}, \mathbf{O}_{\mathrm{A},d+1},\dots, \mathbf{O}_{\mathrm{A},d+k-1}],\\ 
    &\mathbf{Z}_{\mathrm{A},d}^{k} = \textbf{f}_{2}[\mathbf{O}_{\mathrm{A},d}^{k};\theta_{2}],\\
    &\mathbf{Z}_{\mathrm{A}} = \mathcal{G}[\mathbf{Z}_{\mathrm{A},1}^{k}, \mathbf{Z}_{\mathrm{A},2}^{k}, \dots,\mathbf{Z}_{\mathrm{A},D}^{k}].\\
\end{aligned}
\end{equation}

Here, $\mathbf{O}_{\mathrm{A},d}$ is an intermediate feature map for slice $d$ at pre-fusion layer, and $\mathbf{O}_{\mathrm{A},d}^{k}=[\mathbf{O}_{\mathrm{A},d},\mathbf{O}_{\mathrm{A},d+1},\dots,\mathbf{O}_{\mathrm{A},d+k-1}]$. $\textbf{f}_{1}$, $\textbf{f}_{2}$ are two parts of the backbone $\textbf{f}$. This modification helps the network obtain distinguishable features for each slice before fusion, and leads to a better ability to figure out the corresponding relationship between input slices and output slices and a better result. The overall architecture with early-stage multiplexing is shown in Figure~\ref{fig:network_arch}.

\begin{figure}[t]
\begin{center}
  \includegraphics[width=1.0\linewidth]{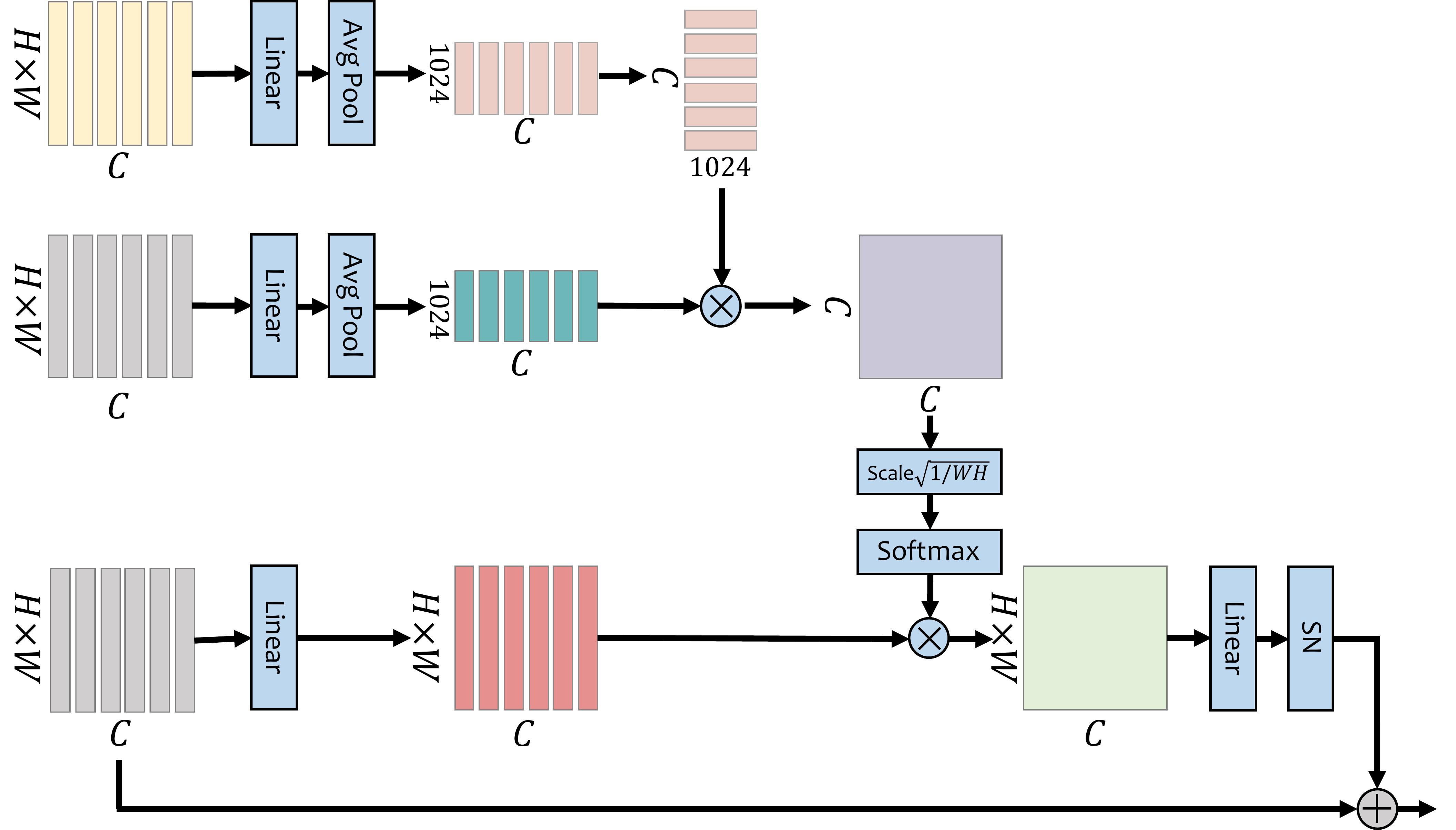}
\end{center}
  \caption{Illustration of the proposed slice-sensitive attention (SSA) module. The design is inspired by non-local module yet focuses on channel-wise relationship instead. (Best viewed in color.)}
\label{fig:SSA_arch}
\end{figure}

\vspace{-0.2cm}
\subsubsection{Slice-Sensitive Attention}
We further address the problem of information loss by introducing slice-sensitive information. In second part of the network, the final feature map is a mixture of inter-slice information and intra-slice information, and the network predicts for different slices based on this same mixed feature map, which is hard if no specific-slice information is given. So we introduce slice-sensitive auxiliary cues from pre-fusion layer to the final feature map. Technically, we add highway connections with a slice-sensitive attention module from pre-fusion layer to prediction layer. And the slice-sensitive information from previous part of the backbone helps make a clearer prediction in attention mechanisms.

Specifically, we introduce slice-sensitive attention (SSA) module, which is illustrated in Figure~\ref{fig:SSA_arch}. We follow~\cite{Wang_2018_CVPR,wu2018long} to design an attention module in non-local manner. However, in our case the main problem lies in confusion in relationship between slices at channel-level. So the attention map should also focus on channel-wise relationship instead of pixel-wise one. Suppose the input feature map has a shape of $(C, H, W)$. Instead of computing the relationship between each location pair, we compute the correspondence between each pair of feature dimension (channel), formulated as $\textbf{y}_{p}=\frac{1}{Z(x)}\sum_{q}f(x_{p},x_{q})g(x_q)$, where $p, q\in\{1,...,C\}$, $x_p, x_q$ represent $H\times W$ dimensional vectors for the $p$-th and $q$-th channel and $Z(x)$ is the normalization coefficient. A pairwise function $f$ computes a scalar between channel $p$ and all channel $q$. The unary function $g$ computes a representation of the input $x$ at the channel $q$. This module intuitively tells where to look at for each dimension of high level feature, thus separate information for each slice from the mixed information during computation.

After equipped with slice-sensitive attention module, the segmentation procedure is updated with:

\begin{equation}
\begin{aligned}
    &\mathbf{U}_{\mathrm{A},d}^{k} = \textbf{f}_{2}[\mathbf{O}_{\mathrm{A},d}^{k};\theta_{2}],\\
    &\mathbf{Z}_{\mathrm{A},d} = \text{SSA}[\mathbf{U}_{\mathrm{A},d}^{k}, \mathbf{O}_{\mathrm{A},d};\eta],\\
\end{aligned}
\end{equation}
where $\text{SSA}$ is a slice-sensitive attention module with parameters $\eta$. $\mathbf{U}_{\mathrm{A},l}^{k}$ is the feature for all k slices, which is the output of second part of the backbone.

\section{Experiments}
\label{Experiments}
\subsection{Datasets and Evaluation}
To verify that our approach can be applied to various organs, we collect a large dataset which contains 200 CT scans. This corpus took 4 full-time radiologists around 3 months to annotate. 
We choose several blood vessels which require more contextual information and also other challenging organs like pancreas and duodenum. We randomly partition the dataset into two parts, one containing 150 cases for training, while another consisting of 50 cases for testing. Each organ is trained and tested individually. When a pixel is predicted as more than one organ, we choose the one with the largest confidence score.

We also conduct experiments on a public dataset from Medical Segmentation Decalthon which contains 303 cases of hepatic vessels, which are challenging if addressed without contextual information. We use 228 cases for training and 75 cases for testing. 

The accuracy of segmentation is evaluated by the Dice-S\o rensen coefficient (DSC): $\mathrm{DSC}$ $(\mathcal{Y},\mathcal{Z}) = \frac{2\times|\mathcal{Y}\cap \mathcal{Z}|}{|\mathcal{Y}|+|\mathcal{Z}|}$. This metric falls in the range of $[0, 1]$, with 1 implying perfect segmentation. We also compare the inference latency of different methods to demonstrate that our method is not only more robust from task to task but also has a much higher inference speed when it is tested in 2D manner.

\subsection{Implementation Details}
\label{train_test_setting}
\subsubsection{Network Architectures}
\vspace{-0.2cm}
We use DeepLabV3+~\cite{chen2018encoder} based on ResNet50~\cite{he2016deep} without ASPP module as our 2D backbone. The input slices are divided into mini-groups, 3 slices in each group. To obtain meaningful intra-slice information, each 3-slice group will go through the first convolution and the following layer1 part of the network. At the fusion stage, we concatenate the feature maps from different groups on channel dimension and use two convolution layers to compress it, one will compress the channel number to its half and another to 256. 

After features go through the remaining part of the backbone as one unity, we apply a slice-sensitive attention module to extract the targeted features for each mini-group. Thus we obtain a prediction focusing on the targeted slices. The slice-sensitive attention module is designed in channel-wise manner because the confusion comes from channel-level instead of pixel-level. Besides, we use an adaptive average pooling to $32\times32$ size followed by a $1\times1$ convolution to pre-process the input. The average pooling and convolution are meant to reduce the variance of different receptive field and also computation cost. We also add a switchable normalization~\cite{luo2018differentiable} at the output to reinforce the module.
\vspace{-0.4cm}

\subsubsection{2D Settings}
\vspace{-0.2cm}
We train the proposed method in a 2D setting. The training phase aims at minimizing DSC loss, which is denoted by ${\mathcal{L}}(Y_{\mathrm{A},d}^{k},P_{\mathrm{A},d}^{k}) = 1-\frac{2\times \sum_{i} Y_{\mathrm{A},d,i}^{k}\cdot P_{\mathrm{A},d,i}^{k}}{\sum_{i} (Y_{\mathrm{A},d,i}^{k}+P_{\mathrm{A},d,i}^{k})}$, where $P_{\mathrm{A},d}^{k}$ is the prediction on the input $k$ slices. The model is initialized with ImageNet~\cite{krizhevsky2012imagenet} pre-trained weight, and trained for 100k iterations with SGD optimizer, where the momentum is 0.9 and weight decay 0.0005. The batch size is 8, and learning rate is 0.005, which decays by a factor of 10 at iteration 70k and 90k. To align different slice sizes so that the model could be trained in parallel, images are cropped or zero padded to make sure that each slice size should be $512\times 512$. Three models on each axis are trained separately. The whole training procedure takes 14 hours on 4 NVIDIA TITAN-Xp GPUs.

At testing phase, the volume is sliced into 15-slice groups where each group has 14 slices overlapped by each other. The final prediction is averaged for each slice. The output ranges in $[0, 1]$ and we use a threshold of 0.5 to get final prediction. Predictions from three viewpoints are fused by majority voting.

\vspace{-0.2cm}
\subsubsection{3D Settings}
\vspace{-0.2cm}
All 3D models are trained with the same 3D setting described in this section, we also train our method in this setting to demonstrate its effectiveness in capturing contextual information. In 3D settings, we adopt SGD optimizer with polynominal learning rate scheduler starting from 0.01 with power of 0.9. The patch size is $128\times128\times64$ and batch size is tuned to take full usage of the 12GB GPU memory. The training procedure lasts for 40k iterations.

At testing stage, we adopt a sliding-window manner with stride $[32, 32, 16]$ for each axis respectively. The probability on each voxel will be averaged based on all probabilities it gets.

\newcommand{\colwidthA}{1.7cm}
\begin{table*}[!tb]
\setlength{\tabcolsep}{0.08cm}
\centering
\begin{tabular}{|l||C{\colwidthA}|C{\colwidthA}|C{\colwidthA}|C{\colwidthA}|C{\colwidthA}|C{\colwidthA}|C{\colwidthA}|}
\hline
Methods                      & Superior m.a.             & Celiac ~~~~a.a.        & Duodenum         & Pancreas         & Vein             & Hepatic Vessel   & Latency ~~~(s)      \\ \hline \hline
3D U-Net~\cite{cciccek20163d}                      & 72.57\%          & 56.31\%          & 68.59\%          & 85.89\%          & 71.05\%          & 55.05\%          & 446.64          \\ \hline
V-Net~\cite{milletari2016v}                        & 72.37\%          & 55.87\%          & 73.18\%          & 85.98\%          & 72.86\%          & 54.38\%          & 833.42          \\ \hline
AH-Net~\cite{liu20183d}                       & 73.03\%          & 56.85\%          & 70.16\%          & 86.40\%          & 71.97\%          & 60.61\%          & 226.14          \\ \hline \hline
\textbf{Ours (3D)}            & 73.82\%          & \textbf{62.66\%} & 70.89\%          & 86.48\%          & 72.36\%          & 59.12\%          & 367.97          \\ \hline
\textbf{Ours (2D) - SingleView} & 73.06\%          & 60.61\%          & 70.16\%          & 86.53\%          & 75.45\%          & 58.33\%          & \textbf{110.75} \\ \hline
\textbf{Ours (2D) - ThreeViews} & \textbf{74.55\%} & 61.61\%          & \textbf{74.71\%} & \textbf{87.36\%} & \textbf{76.23\%} & \textbf{62.67\%} & 335.04          \\ \hline
\end{tabular}
\caption{Experiments on our multi-organ dataset, including superior m.a., celiac a.a., pancreas, duodenum, and vein, and on MSD dataset, including hepatic vessel. All models in 3D settings are trained with patch size $128\times128\times64$ and a customized batch size tuned to take full usage of the GPU memory. Inference time is evaluated with a $512\times512\times394$ case. Inference time of 2D networks with three-views fusion is the summation of three-view inference time. Single-view 2D models are trained and tested on \textit{coronal} view. ``\textbf{Ours (3D)}" means to run our method in the same 3D settings, so that the ability to capture contextual information is fairly comapred between our method and other 3D models. It is noticeable that our method can be run in both 2D and 3D settings, yet 3D models are not capable to run in 2D settings.}
\label{tab:JHMI}
\end{table*}

\subsection{Results on Our Multi-organ Dataset}
\subsubsection{Validation Results}
\vspace{-0.2cm}
Results are summarized in Table~\ref{tab:JHMI}. We have conducted experiments on several challenging blood vessels and organs, including superior m.a., celiac a.a., vein and pancreas, duodenum. On blood vessels, single slice based network usually cannot produce satisfying results, and inter-slice information plays an important role here. We also try to apply our algorithm to other small and challenging organs, proving that our method also works for other more general organs besides blood vessels. With our method, the inter-slice information and intra-slice information collaborate in a better way.

We report the results of our method in 3D setting, 2D single-view setting, and 2D three-views setting respectively. In 3D setting, our model is trained in the same 3D manner as baselines, which is described in~\ref{train_test_setting}. Our model shows a great ability to capture contextual information when trained in the same 3D setting as baselines.

In 2D setting, we report our results based on single-view and three-views fusion, respectively. Our single-view result is good enough and enjoy a lowest inference latency. We also follow typical 2D methods to use a three-view fusion to add contextual information in a post-processing manner, and the results are boosted higher, yet still with a much lower inference latency compared with 3D baselines.

We compare our results with popular 3D models 3D U-Net, V-Net and a hybrid network AH-Net to verify that our model makes good use of 3D information while being efficient, and it turns out our results are consistently better than baselines.
Due to the lack of intra-slice information, a 3D network suffers from serious false positive especially on small blood vessels which only take up thousands of voxels. And because of the lack of pre-trained weight, 3D methods are also unstable on different tasks using the same hyper-parameters. However, our method is more stable and much faster due to the thickened 2D design.

\subsubsection{Diagnoses and Ablation Studies}
We study the proposed Early-Stage Multiplexing (ESM) and Slice-Sensitive Attention (SSA) and different input slice thickness on the multi-organ dataset in an empirical manner.

\newcommand{\colwidthB}{1.3cm}
\begin{table}[tb]
\footnotesize
\setlength{\tabcolsep}{0.04cm}
\centering
\begin{tabular}{|l||C{\colwidthB}|C{\colwidthB}|C{\colwidthB}|C{\colwidthB}|C{\colwidthB}|}
\hline
                         & \begin{tabular}[c]{@{}c@{}}Thickened\\ Inputs\end{tabular} & \begin{tabular}[c]{@{}c@{}}ESM\\ SSA\end{tabular} & \begin{tabular}[c]{@{}c@{}}Superior\\ m.a.\end{tabular}   & \begin{tabular}[c]{@{}c@{}}Celiac\\ a.a.\end{tabular} & Duodenum \\ \hline
\multirow{3}{*}{DeepLab} & $\times$                & $\times$       & 18.60 & 12.79     & 21.88    \\ \cline{2-6} 
                         & $\checkmark$                & $\times$       & 20.70 & 12.99     & 20.68    \\ \cline{2-6} 
                         & $\checkmark$                & $\checkmark$       & \textbf{11.42} & \textbf{11.37}     & \textbf{19.05}    \\ \hline
\end{tabular}
\caption{Inter-slice similarity of different methods, along the \textit{axial} view (smaller numbers are better). Our approach enjoys a better ability of discriminating different slices.}

\label{tab:inter_slice_similarity}
\end{table}

\paragraph{Distinguish-ability. }

We measure the ability of model to distinguish each slice by comparing the prediction of neighboring slices. In detail, we design a statistics, named \textit{inter-slice similarity}, to measure the distinguish-ability of models. We compute the dice score between every pair of neighboring slices along \textit{axial} axis. The closer this number is to inter-slice dice score computed based on ground truth, the better distinguish-ability the model has, which means the prediction has a more similar smoothness as ground truth. We use L2-distance to evaluate the similarity of inter-slice dice score of prediction and that of ground truth. The results are summarized in Table~\ref{tab:inter_slice_similarity}. The distinguish-ability decreased for DeepLab when facing thickened inputs. However, with the proposed method (ESM and SSA), the distinguish-ability increases and is even higher than the model dealing with normal input.

\paragraph{Ways to Introduce Slice-Sensitive Information. }
Based on 12-slice models and superior m.a., we have compared different ways to instantiate attention which combines the specific-slice feature and all-slice feature. As shown in Table~\ref{tab:att_ablation}, directly applying early-stage multiplexing already brings benefits to the model. But the improvement will be less if we introduce slice-sensitive information in an improper way, like concatenate or element-wise dot. With a slice-sensitive module, the feature maps will be re-weighted based on a channel-wise relationship, thus the performance is boosted higher.

\newcommand{\colwidthC}{1.2cm}
\begin{table}[tb]
\small
\setlength{\tabcolsep}{0.06cm}
\centering
\begin{tabular}{|l||c|c|c|c|c|c|}
\hline
            & \begin{tabular}[c]{@{}c@{}}3-slice \\ DeepLab\end{tabular} & \begin{tabular}[c]{@{}c@{}}12-slice \\ DeepLab\end{tabular} & \begin{tabular}[c]{@{}c@{}}12-slice \\ ESM\end{tabular}    & \begin{tabular}[c]{@{}c@{}}12-slice \\ Concat\end{tabular} & \begin{tabular}[c]{@{}c@{}}12-slice \\ Dot\end{tabular} & \begin{tabular}[c]{@{}c@{}}12-slice \\ SSA\end{tabular} \\ \hline
DSC$_{\mathrm{C}}$      & 71.20\%          & 70.15\%           & 72.73\%          & 72.27\%         & 71.16\%      & \textbf{73.23\%}           \\ \hline
DSC$_{\mathrm{S}}$      & 70.71\%          & 67.03\%           & \textbf{71.91\%} & 71.83\%         & 71.75\%      & 71.67\%                    \\ \hline
DSC$_{\mathrm{A}}$      & 69.62\%          & 70.35\%           & 72.18\%          & 72.04\%         & 71.21\%      & \textbf{72.54\%}           \\ \hline
DSC$_{\mathrm{F}}$ & 73.32\%          & 71.07\%           & 73.63\%          & 73.07\%         & 73.74\%      & \textbf{74.17\%}           \\ \hline
\end{tabular}
\caption{Compare different settings on superior m.a.. DeepLab serves as 2D backbone for all methods. ESM stands for Early-Stage Multiplexing. Concat, dot, and SSA are based on the ESM model and take different way to introduce slice-sensitive information to the final feature map. The subscripts $\mathrm{C}$, $\mathrm{S}$, $\mathrm{A}$, $\mathrm{F}$ represents plane \textit{Coronal}, \textit{Sagittal}, \textit{Axial}, and \textit{Fusion} respectively.}

\label{tab:att_ablation}
\end{table}

\newcommand{\colwidthD}{1.5cm}
\begin{table*}[!tb]
\setlength{\tabcolsep}{0.12cm}
\centering
\begin{tabular}{|l||C{\colwidthD}|C{\colwidthD}|C{\colwidthD}|C{\colwidthD}|C{\colwidthD}|C{\colwidthD}|C{\colwidthD}|C{\colwidthD}|}
\hline
            & 3-slice & 6-slice & 9-slice     & 12-slice    & 15-slice   & 18-slice & 21-slice    & 24-slice \\ \hline
DSC$_{\mathrm{C}}$      & 70.37\%      & 72.49\%      & 73.07\%          & \textbf{73.23\%} & 73.06\%          & 73.04\%       & 72.73\%          & 72.72\%       \\ \hline
DSC$_{\mathrm{S}}$      & 70.49\%      & 71.34\%      & 71.65\%          & 71.67\%          & 72.97\%          & 72.17\%       & \textbf{73.17\%} & 72.46\%       \\ \hline
DSC$_{\mathrm{A}}$      & 70.70\%      & 70.59\%      & \textbf{72.76\%} & 72.54\%          & 72.64\%          & -             & -                & -             \\ \hline
DSC$_{\mathrm{F}}$ & 72.61\%      & 73.21\%      & 74.05\%          & 74.17\%          & \textbf{74.55\%} & 73.61\%       & 74.17\%          & 73.78\%       \\ \hline
\end{tabular}
\caption{Dice score of our method (Early-Stage Multiplexing + Slice-Sensitive Attention) with different slice thickness on superior m.a.. We try different slice thickness from 3 to 24. It can be observed that our model performs well even for a much larger slice thickness. '-' means the model does not converge, if some plane result is missing, the fusion will be the average on the remaining planes.}
\label{tab:thickness_ablation}
\end{table*}

\paragraph{Slice Thickness. }
We try different slice thickness to test our method capacity, from 3-slice to at most 24-slice with same batch size under 12G GPU memory limitation. The results are summarized in Table~\ref{tab:thickness_ablation}. We find that the results keep increasing until slice thickness reaches 15, while the \textit{axial} model with over 18-slice model does not converge. The trend of performance increase is different for each viewpoint when slice thickness increases. For plane \textit{coronal}, a major improvement happens when slice thickness increase from 3 to 6 (+2.12\%). And models with slice thickness over 9 produce similar results. For plane \textit{sagittal}, there happen two major improvements, one is from 3 to 6 (+0.85\%) and another is from 12 to 15 (+1.30\%), while 6-slice, 9-slice, and 12-slice have similar accuracy. For plane \textit{axial}, when increasing slice numbers from 6 to 9, the result increases by 2.17\%. It proves that there exist some bottlenecks when increasing slice thickness, and breaking through the bottleneck brings most significant improvement.

We can also observe that with thickened inputs, single-axis result is closer to the three-axes fusion one, which indicates that our method does bring contextual information into the network, yet unlike three-view fusion which is in a post-processing manner, our method works at both training and testing stages.

\subsection{Results on the MSD Dataset}
We further apply our approach on a public dataset -- hepatic vessel segmentation in Medical Segmentation Decathlon~\cite{simpson2019large}. Hepatic vessels have a tree-structure and are challenging if addressed by common 2D networks. Results are shown in Table~\ref{tab:JHMI}. Our results outperform competitors, which illustrates the ability of our approach in capturing contextual information for such a tree-structure target.

\begin{figure}[!tb]
\begin{center}

  \includegraphics[height=1.0\linewidth]{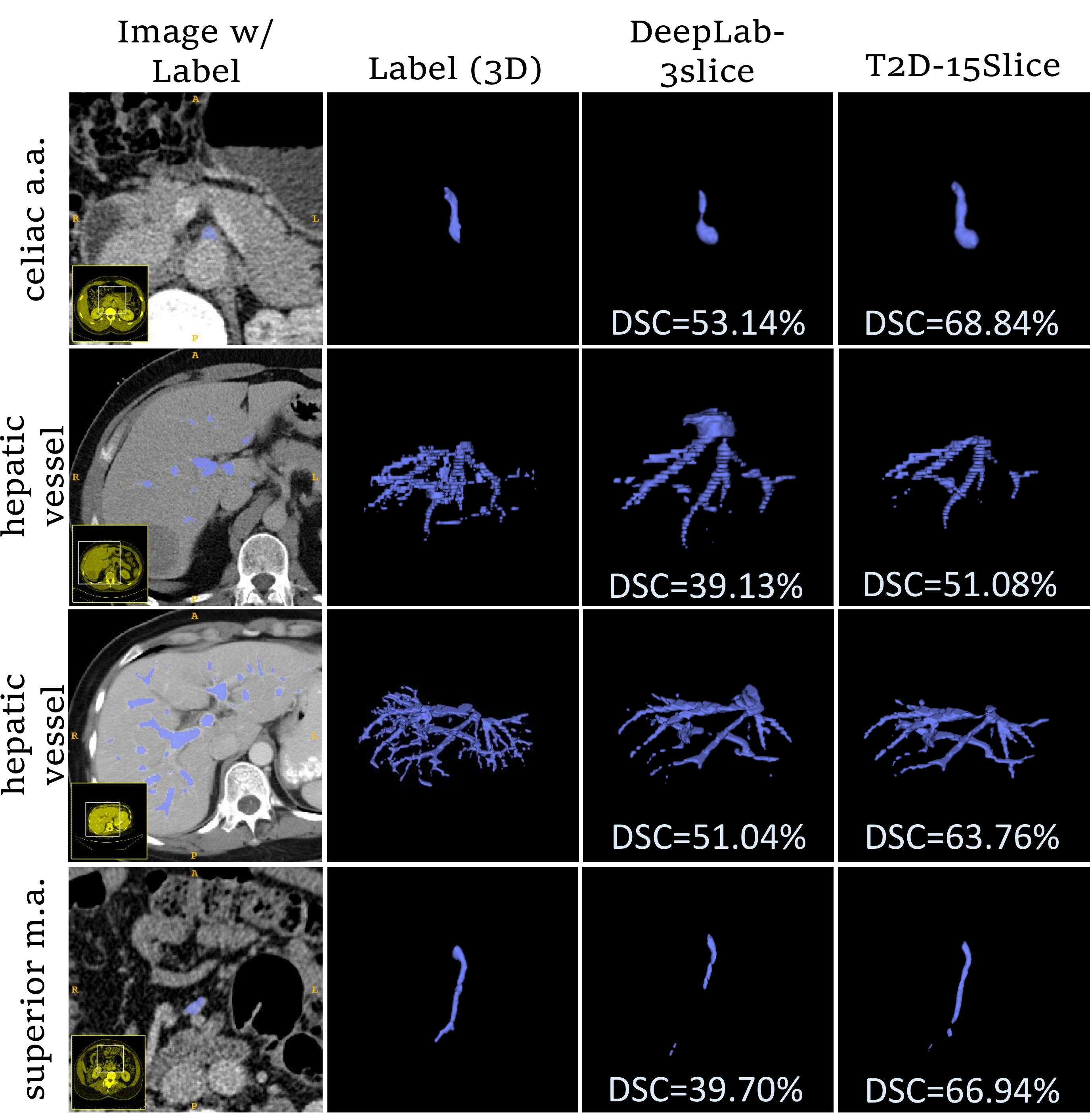}
\end{center}

  \caption{3D visualization comparison for celiac a.a., hepatic vessel, and superior m.a., respectively. Compared with baseline 3-slice DeepLab, our method shows a greater ability in capturing the continuous attributes of blood vessels. (Best viewed in color.)}
\label{fig:vis1}

\end{figure}

We visualize two cases of hepatic vessel and one case of celiac a.a. and superior m.a. with ITK-SNAP~\cite{py06nimg} (see Fig~\ref{fig:vis1}). First two rows are for hepatic vessel, which is contiguous and like a multi-branch tree. Compared with baseline DeepLab, our method with thickened inputs does better in capturing this vessels' contiguity. In the fourth row of superior m.a., DeepLab fails to predict at some points, resulting in a worse continuity of the prediction. Yet our method makes accurate prediction at thin bottlenecks of blood vessel with captured contextual information. We observe no obvious failure (a typical fail in baseline single axis result) and a better segmentation consistency in our approach.

\section{Conclusions}
\label{Conclusions}

This paper is motivated by the need of capturing 3D contexts, and aims at designing a network structure which works on thickened 2D inputs. The major obstacle is the information loss brought by fusion along the third dimension. We propose two keys to deal with this issue, {\em i.e.}, postponing the stage of information fusion by early-stage multiplexing, and creating a shortcut connection based on slice-sensitive attention module between the pre-fusion stage and the final decision stage. Evaluated on a few medical image segmentation datasets, our approach reports higher segmentation accuracy and lower inference latency with thickened input data, demonstrating the effectiveness of capturing 3D contexts.

Our research sheds light on designing efficient 3D networks for segmenting volumetric data. The success of our approach provides a piece of side evidence that both 2D and 3D networks are not the optimal solution, as 2D methods benefit from natural image pre-training but inevitably lack contexts, meanwhile 3D methods usually waste time and memory for unnecessary computations. Absorbing advantages from both of them remains an open problem and deserves more efforts in future research.

{\small
\bibliographystyle{ieee_fullname}
\bibliography{egbib}
}

\end{document}